\documentclass[lettersize, screen, journal]{IEEEtran}
\usepackage{amsmath}
\usepackage{amsfonts}
\usepackage{array}
\usepackage[caption=false, font=normalsize, labelfont=sf, textfont=sf]{subfig}
\usepackage{textcomp}
\usepackage[ruled]{algorithm2e}
\usepackage{setspace}
\usepackage{colortbl}
\definecolor{shadegray}{RGB}{230,230,230}
\definecolor{lightgray}{RGB}{230,230,230}
\usepackage{float}
\usepackage[table]{xcolor}
\usepackage{mathrsfs}
\usepackage{nicefrac}
\usepackage{multirow}
\usepackage{makecell}
\usepackage{booktabs}
\usepackage{bbding}
\usepackage[pagebackref, breaklinks, colorlinks,
citecolor=green,
linkcolor=red,
urlcolor=blue]{hyperref}

\newcommand{\lredcell}{\color{red!60!black}}
\newcommand{\lgreencell}{\color{green!50!black}}
\newcommand{\lbluecell}{\color{blue!60!black}}

\newcommand{\gr}[1]{\cellcolor{shadegray}#1}

\usepackage{bm}
\usepackage{stfloats}
\usepackage{url}
\usepackage{verbatim}
\usepackage{graphicx}
\usepackage{cite}
\usepackage{tikz}
\usepackage[numbers, sort&compress]{natbib}
\begin{document}
	
	\title{Image Quality Assessment: Exploring Quality Awareness via Memory-driven Distortion Patterns Matching}
	
		\author{\IEEEmembership{}
		Xuting Lan, Mingliang Zhou, \IEEEmembership{Senior Member, IEEE}, Xuekai Wei,
		Jielu Yan, Yueting Huang, Huayan Pu, Jun Luo, and Weijia Jia, \IEEEmembership{Fellow, IEEE}~
}

	\markboth{}%
	{Shell \MakeLowercase{\textit{et al.}}: A Sample Article Using IEEEtran.cls for IEEE Journals}
	
	
	\maketitle

\begin{abstract}
Existing full-reference image quality assessment (FR-IQA) methods achieve high-precision evaluation by analysing feature differences between reference and distorted images. However, their performance is constrained by the quality of the reference image, which limits real-world applications where ideal reference sources are unavailable. Notably, the human visual system has the ability to accumulate visual memory, allowing image quality assessment on the basis of long-term memory storage. Inspired by this biological memory mechanism, we propose a memory-driven quality-aware framework (MQAF), which establishes a memory bank for storing distortion patterns and dynamically switches between dual-mode quality assessment strategies to reduce reliance on high-quality reference images. When reference images are available, MQAF obtains reference-guided quality scores by adaptively weighting reference information and comparing the distorted image with stored distortion patterns in the memory bank. When the reference image is absent, the framework relies on distortion patterns in the memory bank to infer image quality, enabling no-reference quality assessment (NR-IQA). The experimental results show that our method outperforms state-of-the-art approaches across multiple datasets while adapting to both no-reference and full-reference tasks. 
\end{abstract}

\section{Introduction}
In the field of digital image processing and multimedia applications, image quality assessment (IQA) is a fundamental task, as it not only directly affects the end user's viewing experience but also impacts subsequent processing and analysis workflows \cite{xian, zheng2021learning, shen, liao2025image, transformer4}. Full-reference image quality assessment (FR-IQA) \cite{liao2024image, zhou2024hdiqa, lang} remains one of the most widely used approaches, in which the visual quality of a distorted image is evaluated by comparing it with a pristine reference image. Although traditional metrics such as mean squared error (MSE), mean absolute error (MAE), and peak signal-to-noise ratio (PSNR) are easy to compute, they often do not align well with human visual perception. To bridge this gap, deep learning-based full-reference image quality assessment methods have emerged as powerful alternatives, leveraging pre-trained convolutional neural networks (CNNs) \cite{VGG, resnet50} to extract hierarchical semantic features from both reference and distorted images \cite{Swiniqa, AHIQ, Musiq, Topiq}. Ding \textit{et al.} proposed a method called DISTS \cite{DISTS}, which captures structural and textural differences for image quality assessment. Building upon this, A-DISTS \cite{ADISTS} further improves performance by adaptively modeling local distortions. In addition, Zhang \textit{et al.} introduced LPIPS \cite{LPIPS}, and Liao  \textit{et al.} developed DeepWSD \cite{liao2022deepwsd}, both of which compare images in deep feature space to better reflect perceptual similarity. Despite their success, these methods fundamentally rely on the availability of high-quality reference images. In real-world scenarios, however, such reference images are often degraded or entirely unavailable, which significantly limits the practical utility of full-reference IQA models. When the reference image itself is slightly distorted or missing, the evaluation results can become unreliable or even misleading, undermining the theoretical advantages of FR-IQA. Nevertheless, we do not directly shift to no-reference IQA (NR-IQA) \cite{zhang,WaDIQaM-FR,Topiq} but instead continue to explore within the FR-IQA framework. On the one hand, FR-IQA still provides stable and fine-grained supervision when reference images are available, which facilitates model learning. On the other hand, enhancing its robustness under nonideal reference conditions not only broadens its practical applicability but also lays the groundwork for building a unified framework compatible with NR-IQA.

\begin{figure}[!t]
	\centering
	\hspace{5mm}
	\setlength{\abovecaptionskip}{0cm}
	\includegraphics[width=0.38\textwidth]{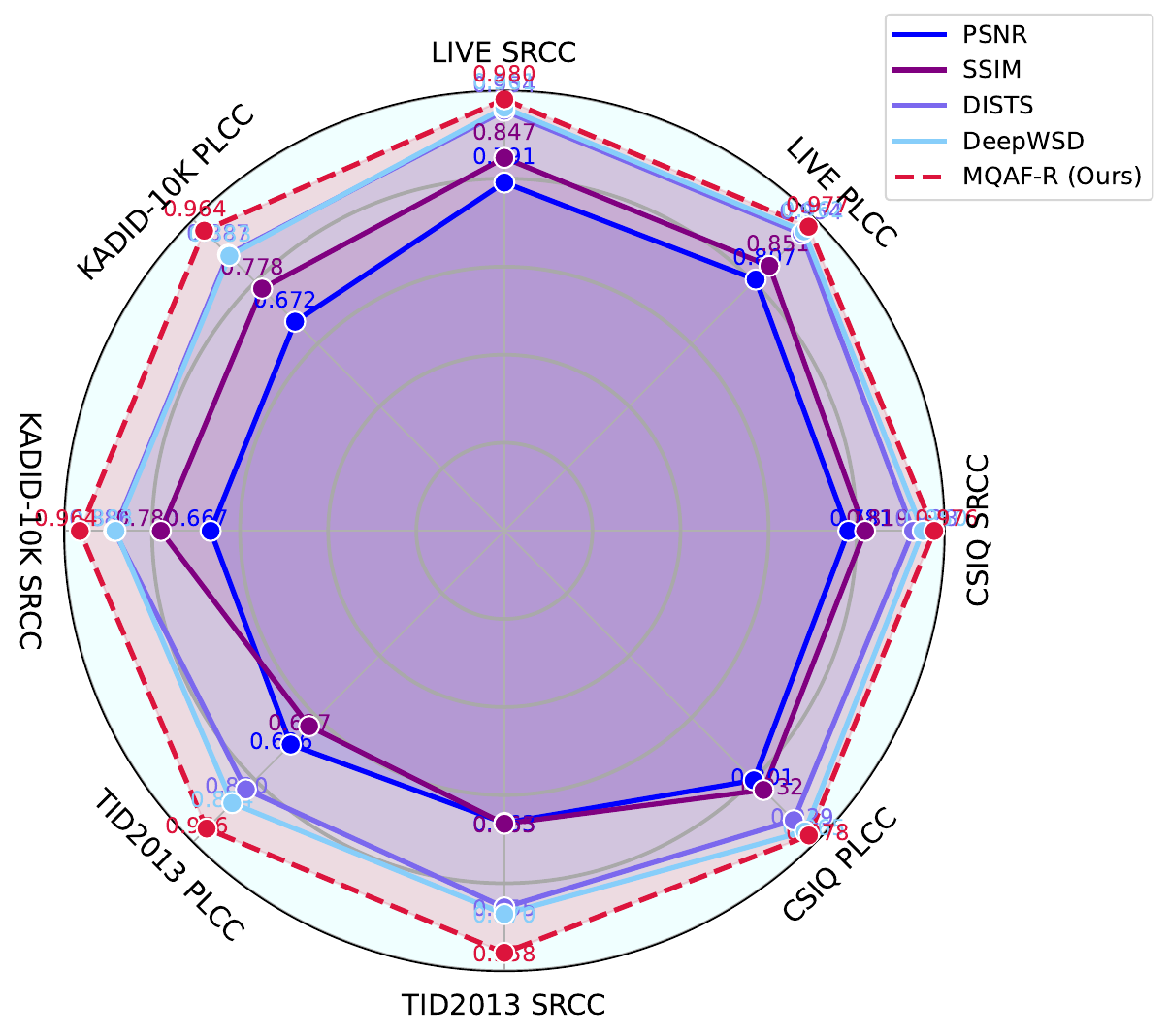}
	\caption{Comparison of MQAF with existing IQA methods, highlighting the exceptional IQA ability of MQAF-R.}
	\label{radar}
\end{figure}
To address this limitation, we propose a memory-driven quality-aware framework (MQAF) that aims to reduce the dependence on high-quality reference images by introducing a learnable memory mechanism that stores distortion patterns. This design is inspired by findings in cognitive psychology \cite{tversky1974judgment, stevens1957psychophysical,marks2014sensory}: humans often assess the quality of new stimuli by comparing them to previously encountered examples stored in memory, known as standard stimuli or reference anchors. For example, when rating the quality of a new image, an observer intuitively relates it to similar images they have seen before and adjusts the perceived quality depending on whether the current distortion appears more or less severe. This reference process highlights the potential of explicit memory-based representations in enhancing perceptual evaluation, particularly in the absence of reliable reference images. 
Stevens’ power law \cite{ stevens1957psychophysical} explains that the subjective perception of stimulus intensity follows a nonlinear function of its physical magnitude, suggesting that perceptual judgments are not based solely on current inputs but also shaped by internalized standards. Similarly, Tversky and Kahneman’s anchoring and adjustment heuristic \cite{tversky1974judgment} shows that people make judgments by adjusting from prior anchors formed from previous experience, especially under uncertainty. In the context of image quality assessment, this implies that observers rely on memory-based comparisons rather than absolute references. 
 On the basis of this insight, MQAF constructs a memory bank that stores distortion patterns extracted from past examples and leverages them to flexibly guide image quality assessment under both full-reference and no-reference conditions. Unlike conventional deep models that implicitly memorize data distributions through network parameters, the memory bank in MQAF is an explicitly accessible and updatable structure, which enables the model to actively retrieve distortion patterns similar to the current input during inference. This mechanism not only enhances the model’s adaptability to previously unseen distortion types but also improves interpretability of the assessment process, overcoming the generalization limitations inherent in purely data-driven approaches.

\begin{itemize}
	\item We propose a novel memory-driven quality-aware assessment framework (MQAF) that constructs a trainable memory bank composed of memory units to achieve autonomous learning and storage of typical distortion pattern features. This memory bank is able to continuously update and store new distortion types while maintaining long-term memory and precise recognition of key distortion patterns. This fundamentally overcomes the traditional method’s dependence on high-quality reference images and provides a novel solution to the problem of obtaining reference images in practical applications.
	\item The MQAF framework we designed exhibits excellent versatility and flexibility and is capable of adapting to both reference and no-reference scenarios. When a reference image is available, MQAF can combine the reference information with distortion patterns in the memory bank for comprehensive adaptive evaluation, thereby improving the accuracy of quality prediction. When the reference image is unavailable, MQAF can rely fully on the distortion patterns in the memory bank for quality scoring, enabling NR-IQA. This dual-mode operation mechanism allows MQAF to function effectively in various real-world applications, expanding the scope of image quality assessment.
	\item The experimental results demonstrate that our method outperforms existing state-of-the-art methods across multiple public datasets and evaluation metrics. Particularly in no-reference scenarios or when the reference image quality is low, MQAF maintains a stable performance advantage, showing exceptional robustness and broad application potential. These experimental results not only validate the effectiveness of the memory mechanism in image quality assessment but also provide new research directions for the development of future image quality assessment technologies.
\end{itemize}

The structure of this paper is as follows. In \autoref{sec:formatting}, we review the existing work related to IQA methods. In \autoref{Formulation}, we describe the problem formulation of IQA. In \autoref{Methodology}, we provide a detailed description of the proposed MQAF method. In \autoref{Experiments}, we present the experimental setup and analyse the results. Finally, the \autoref{Conclusion} concludes the paper.
\section{Related Work}
\label{sec:formatting}

\subsection{Image Quality Assessment}
In the field of IQA, various classical FR metrics have been developed to simulate HVS perception of image distortions. Among them, Wang \textit{et al.} proposed the structural similarity index (SSIM) \cite{wang2004image}, which calculates image similarity on the basis of luminance, contrast, and structural information, serving as the foundation for many subsequent methods. They later introduced the multi-scale SSIM (MS-SSIM) \cite{MS-SSIM}, which evaluates image quality across multiple scales. Wu \textit{et al.} \cite{VIF} proposed the visual information fidelity (VIF) index from an information-theoretic perspective, leveraging natural scene statistics to measure the amount of visual information retained in distorted images. Larson \textit{et al.} \cite{CSIQMAD} introduced the most apparent distortion (MAD) model, which combines assessments of both subtle and obvious distortions for a consistent evaluation. Zhang \textit{et al.} \cite{FSIM} proposed the feature similarity index (FSIM), which focuses on perceptual features via phase congruency and gradient magnitude. They further introduced the visual saliency-induced index (VSI) \cite{VSI}, emphasizing the influence of visually salient regions on quality assessment. These methods involve advanced image quality evaluation, providing multidimensional analytical tools for image processing and compression.
\subsection{Deep feature-based image quality assessment}
In recent years, deep learning methods have been extensively applied to FR-IQA, achieving breakthroughs in perceptual assessment through distance-based feature similarity metrics. Wang \textit{et al.} \cite{wang2005reduced} pioneered the use of Kullback-Leibler divergence (KLD) to measure histogram similarity for reduced-reference IQA (RR-IQA), effectively quantifying distortion levels by capturing distribution differences between probability distributions and simulating the sensitivity of HVS to image information changes. Compared with KLD, Liu \textit{et al.} \cite{liu2016perceptual} introduced Jensen–Shannon divergence (JSD) to address contrast distortion evaluation, demonstrating superior robustness in handling significant distribution disparities and resolving asymmetry issues. With advancements in deep learning, Zhang \textit{et al.} \cite{LPIPS} proposed the LPIPS metric, which leverages deep neural networks (DNNs) to extract image features and predict quality scores through feature distribution comparisons. Delbracio \textit{et al.} \cite{delbracio2021projected} expanded distribution-based metrics by projecting image features into high-dimensional spaces and utilizing Wasserstein distance (WSD) to quantify distribution discrepancies between enhanced and target images. Building on this, Liao \textit{et al.} \cite{liao2022deepwsd} integrated WSD into FR-IQA frameworks, achieving robust quality prediction by matching distributions in deep feature spaces. To further advance the field, Liao \textit{et al.} \cite{liao2024image} developed a training-free FR-IQA framework incorporating multiple perceptual distance metrics (JSD, WSD, KLD) to compare deep feature distributions, demonstrating the broad applicability of distribution-based metrics in IQA tasks. In addition, some no-reference image quality assessment (NR-IQA) methods \cite{gan2,maeiqa, vcrnet} generate pseudoreference images by restoring distorted images as auxiliary information, thereby quantifying quality differences. These frameworks provide flexible solutions that can adapt to various distortion types and scenarios, significantly advancing the development of IQA methods.

These methods rely heavily on high-quality reference images as evaluation benchmarks, with their core idea being to quantify image quality by computing feature differences between the reference and distorted images. However, in practical applications, high-quality reference images are often difficult to obtain, especially in scenarios such as image transmission, storage degradation, or content enhancement, where the reference image may be missing or already degraded. This leads to complete failure or weakened effectiveness of the assessment, significantly limiting the applicability of traditional FR-IQA methods. Therefore, a key challenge for current FR-IQA methods is how to reduce the dependence on reference images while still achieving accurate IQA.
\section{Problem Formulation}
\label{Formulation}
Traditional FR-IQA methods typically rely on a pristine reference image as a benchmark, with the optimization objective formulated as: 
\begin{equation}
	\begin{aligned}
		\min_{\theta, \zeta} \mathbb{E} \left[ l\left( \text{Dis}_\zeta(\phi_\theta(I_\text{ref}), \phi_\theta(I_\text{dist})), Q^{\text{true}} \right) \right]
	\end{aligned}
\end{equation}
where \( I_\text{ref} \) and \( I_\text{dist} \) denote the reference and distorted images, respectively, \(\phi_\theta(\cdot)\) is a feature extraction function parameterized by \(\theta\). \(\text{Dis}_\zeta(\cdot, \cdot)\) represents a distance metric controlled by the parameter \(\zeta\), and $Q^{\text{true}}$ denotes the ground truth score. However, this approach heavily depends on an undistorted reference image, which may be unavailable or incomplete in real-world applications. This limitation reduces the applicability of traditional full-reference methods, affecting the stability and generalization capability of quality assessment. In contrast, humans can perceive image quality even in the absence of a reference image by relying on prior visual experiences. {This capability arises from the continuous storage and accumulation of various distortion patterns in the human brain over time, utilizing an internal memory system to retrieve, update, and compare these patterns for quality evaluation.} Inspired by this memory mechanism, we propose a memory-driven quality-aware assessment framework (MQAF), which builds a {memory bank} to store and learn different types of distortion patterns, thereby enhancing quality awareness. The memory storage and retrieval process is formulated as: 
\begin{equation}
	\begin{aligned}
	 \mathcal{M}_\omega(\phi_\theta(I_\text{dist}), \mathcal{D}_k)
	\end{aligned}
\end{equation}
where \(\mathcal{M}_\omega(\cdot)\) represents the memory storage and retrieval mechanism, \(\omega\) is the parameter controlling memory storage and updating, and \(\mathcal{D}_k\) denotes the \( k \)-th distortion pattern stored in the {memory bank}. Through this mechanism, the model can dynamically switch between different quality assessment modes, enabling effective evaluation in both full-reference (FR) and no-reference (NR) scenarios. This reduces reliance on reference images and enhances the robustness of the assessment process. 

\begin{figure*}[!t]
	\centering
	\setlength{\abovecaptionskip}{0.cm}
	\includegraphics[width=0.95\textwidth]{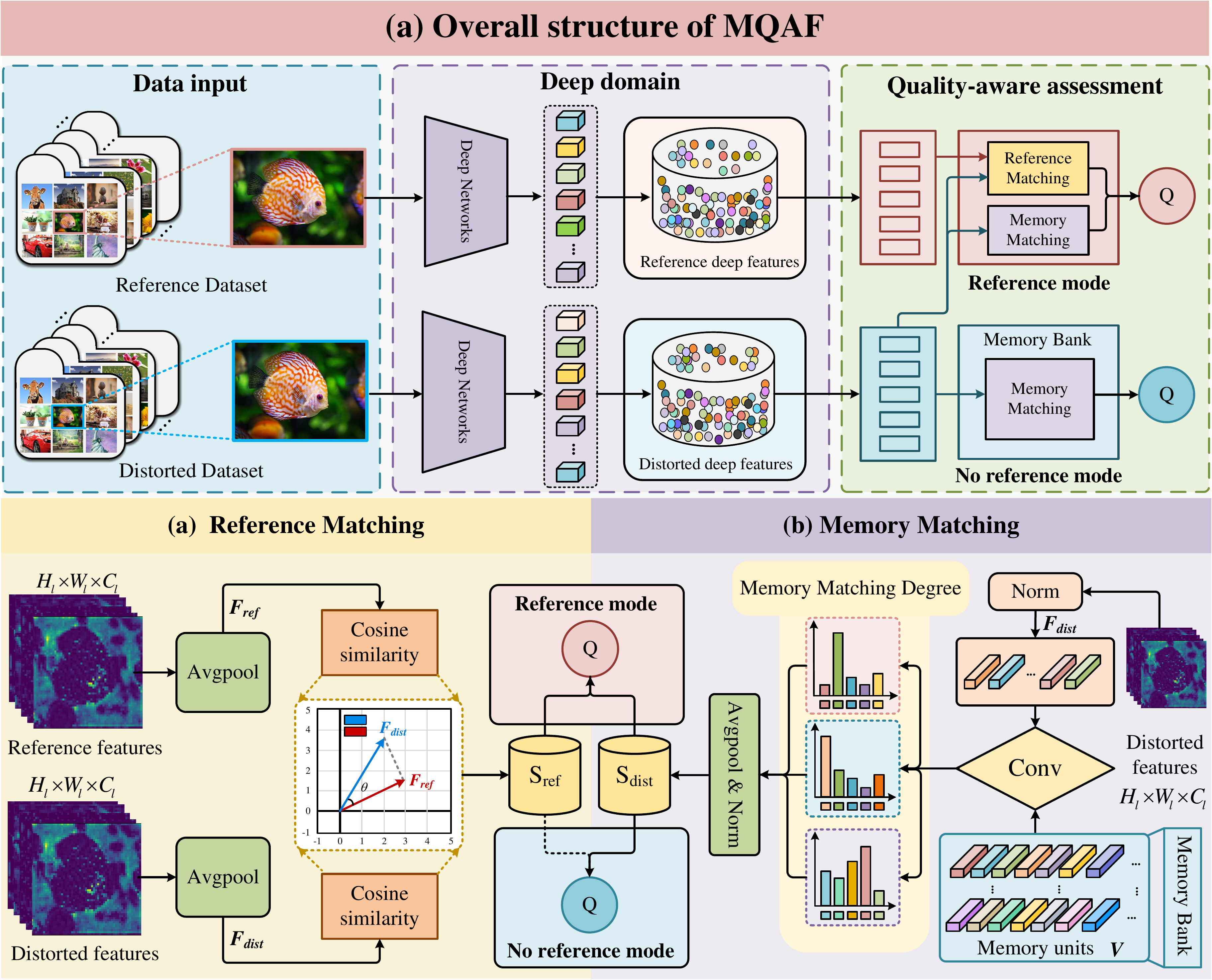}
	\vspace{0.1em}
	\caption{{The framework (MQAF) of our proposed method can adaptively select different modes based on the availability of reference images in the quality assessment process. When a reference image is available, the framework enters the reference mode, where the final quality score \( Q \) is obtained by computing reference matching and memory matching. When no reference image is available, the framework switches to the no-reference mode, where the final quality score \( Q \) is determined by matching with distortion patterns (memory units \(\mathcal{V}\)) stored in the memory bank.}}
	\label{method}
\end{figure*}
Based on the established memory mechanism, we employ an end-to-end learning approach to minimize the loss function \( l(\cdot) \) over the entire dataset, ensuring that the predicted quality score \( Q \) closely approximates the ground truth quality score \( Q^{\text{true}} \). Our optimization objective is defined as follows:
\begin{equation}\label{LOSS}
	\begin{aligned}
		\min_{\theta, \zeta,\omega} \mathbb{E}[l( \text{Dis}_\zeta(\phi_\theta(I_\text{ref}), \phi_\theta(I_\text{dist}) + \mathcal{M}_\omega(\phi_\theta(I_\text{ref})) ), Q^{\text{true}})]
	\end{aligned}
\end{equation}
where \(\theta\) is optimized to refine the feature extractor \(\phi_\theta(\cdot)\), ensuring effective representation of image quality; \(\zeta\) is optimized to enhance the distortion metric \(\text{Dis}_\zeta(\cdot, \cdot)\), which quantifies quality differences between images; and \(\omega\) is trained to improve the memory module \(\mathcal{M}_\omega(\cdot)\), enabling efficient storage and retrieval of learned distortion patterns. The {memory bank} \(\mathcal{M}_\omega(\cdot)\) not only continuously stores and updates distortion types but also maintains the independence of different distortion patterns. This prevents redundant interference, reinforces long-term memory for various distortions, and ultimately allows the model to assess image quality effectively even in the absence of a reference image.
\section{Methodology}\label{Methodology}
\subsection{Framework}
To address the overreliance on high-quality reference images in IQA, we propose a memory-driven quality-aware assessment framework (MQAF), as shown in Figure \ref{method}. The core of this framework is a trainable memory bank composed of memory units, which stores and extracts typical distortion pattern features. The memory bank not only retains long-term memories of key distortion patterns but also adapts to new distortion types, thus overcoming the dependence on high-quality reference images found in traditional methods. {Initially}, deep features of both reference and distorted images are extracted via a deep network and serve as the foundation for distortion pattern matching on the basis of the memory mechanism. Specifically, we use a ResNet50 pretrained model as the encoder, with reference and distorted images \(I_{\text{ref}}, I_{\text{dist}} \in \mathbb{R}^{N \times H \times W \times 3}\), where \(N\) is the sample size; \(H\) and \(W\) are the image height and width, respectively; and 3 represents the RGB channels. {We extract high-level features from the final layer, followed by normalization and global average pooling to obtain the reference and distorted features \(F_{\text{ref}}, F_{ \text{dist}} \in \mathbb{R}^{2048}\).} In the quality assessment phase, the framework adjusts the evaluation mode on the basis of the availability of the reference image.

When the reference image is available, the MQAF first measures the reference matching score between $F_{ \text{ref}}$ and $F_{ \text{dist}}$. Then, $F_{ \text{dist}}$ are matched with the distortion patterns stored in the memory bank to obtain the memory matching score. Finally, the final quality score is obtained through adaptive weighting. When the reference image is unavailable, the MQAF completely relies on the distortion patterns in the memory bank for quality assessment, calculating the final quality score by matching the high-level features of the distorted image with the stored distortion patterns. During the training process, the memory loss continuously optimizes the distortion patterns stored in the memory bank, enabling the model to effectively learn and memorize different types of distortions, improving its adaptability to unknown distortion types. This ensures that the model can achieve robust quality assessment regardless of the availability of reference images.

\subsection{Memory-driven Distortion Pattern Storage and Matching Metric}
	In human visual perception, the brain continuously stores and updates information in response to new stimuli, thereby forming memory. Inspired by this biological mechanism, we propose a memory-driven quality assessment framework (MQAF) that simulates the processes of memory storage and updating (during training) and memory retrieval (during inference) to increase the adaptability and accuracy of IQA under varying reference conditions. To this end, we construct a trainable memory bank composed of multiple memory units, which is progressively optimized during training to learn prototypical representations of various distortion types. Each memory unit, as a fundamental component of the memory bank, represents a prototype of a distortion pattern, that is, the characteristic feature distribution exhibited by images under different types of degradation. Specifically, the memory bank is defined as $\mathcal{V} \in \mathbb{R}^{256 \times 2048}$, where 256 memory units are stored, each represented by a 2048-dimensional feature vector. MQAF dynamically switches its evaluation strategy on the basis of the availability of reference images, enabling more flexible and accurate quality prediction. The quality score \(Q\) is calculated as follows:
\begin{equation}
\label{Q_score}
	\begin{aligned}
		Q = \alpha S_{\text{ref}} + (1 - \alpha) S_{\text{dist}}
	\end{aligned}
\end{equation}
	where \(S_{\text{ref}}\) denotes the reference matching score. \(S_{\text{dist}} \) represents the memory matching score. \(\alpha \) is a learnable weight parameter within the range [0,1] used to balance the contributions of the reference matching score and the memory matching score.

When the reference image is available, the quality score $Q$ is obtained through adaptive weighting of the reference matching score \(S_{\text{ref}}\) and the memory matching score \(S_{\text{dist}}\). First, on the basis of the reference features $F_{\text{ref}}$ and distorted features $F_{\text{dist}}$, we compute the matching score via the cosine similarity $S_{\text{cos}}$ and norm-based similarity $S_{\text{norm}}$, to obtain the final reference matching score $S_{\text{ref}}$. Using only cosine similarity ignores the magnitude information of the features, as it focuses solely on the angle between feature vectors while neglecting their length (magnitude). In practice, two images may have features pointing in similar directions (i.e., high cosine similarity) but differ significantly in magnitude, which may indicate severe distortion in perceptual quality. Therefore, by combining both directional and magnitude information, the matching between the reference and distorted images in the feature space can be assessed more comprehensively.
\begin{equation}
	\begin{aligned}
		S_{\text{ref}} = \text{Distance}(F_{\text{ref}}, F_{\text{dist}}) = S_{\text{cos}} \cdot S_{\text{norm}} \\
		\left\{
		\begin{aligned}
			S_{\text{cos}} &= \frac{F_{\text{ref}} \cdot F_{\text{dist}}}{\|F_{\text{ref}}\|_2 \cdot \|F_{\text{dist}}\|_2} ,\\
			S_{\text{norm}} &= 1 - \frac{\left| \|F_{\text{ref}}\|_2 - \|F_{\text{dist}}\|_2 \right|}{\max(\|F_{\text{ref}}\|_2, \|F_{\text{dist}}\|_2)}
		\end{aligned}
		\right.
	\end{aligned}
\end{equation}
	Next, the memory matching score \(S_{\text{dist}}\) between the distorted image and the distortion patterns stored in the memory bank is computed. Specifically, {the distortion features \(F_{\text{dist}}\)} are matched with all distortion patterns in the memory bank through a convolutional projection, obtaining local distortion pattern matching results. Then, global average pooling is applied to aggregate the full-image matching results, producing a global matching feature vector. The L2 norm of this vector is computed to measure the overall matching degree between the input distorted image and the stored distortion patterns. The final memory matching score \(S_{\text{dist}}\) is calculated as follows:
	\begin{equation}
	\begin{aligned}
		\small
		S_{\text{dist}} = \left\| \text{Pool} \left( \text{Conv} \left( \frac{F_\text{dist}}{\|F_\text{dist}\|}, \frac{\mathcal{V}}{\|\mathcal{V}\|} \right) \right) \right\|
	\end{aligned}
\end{equation}
where \(\text{Conv}(\cdot, \cdot) \) denotes the matching operation using the memory bank as the convolutional kernel, \(\text{Pool}(\cdot) \) refers to global average pooling, and \(\|\cdot\| \) represents L2 normalization. Specifically, \(\|\cdot\| \) computes the L2 norm of the feature vector, which measures the matching degree between the distorted image and the stored patterns. Finally, we dynamically compute the weights \(\alpha\) of the reference and memory matching scores via a simple adaptive weight network (AWN). To guide the learning of \(\alpha\), we supervise it with an error-based confidence loss.
\begin{equation}
	\begin{aligned}
		\mathcal{L}_{\alpha} &= (\alpha - \alpha_{\text{target}})^2
	\end{aligned}
\end{equation}
\begin{equation}
	\small
	\left\{
	\begin{aligned}
		\alpha &= \text{AWN} \left(\text{concat} \left(\text{Pool} \left(F_\text{dist} \right),\ \text{Pool} \left(F_\text{ref} \right) \right) \right) \text{,}\\
		\alpha_{\text{target}} &= \frac{e^{|S_{\text{dist}} - Q^{\text{true}}|}}{e^{| S_{\text{dist}} - Q^{\text{true}}|} + e^{| S_{\text{ref}} - Q^{\text{true}}|}} 
	\end{aligned}
	\right.
\end{equation}	
where $\text{concat}(\cdot)$ denotes the concatenation operation, $\alpha_{\text{target}}$ denotes the target weight, and \(\mathcal{L}_{\alpha}\) reflects the error ratio of the two scores \(S_{\text{ref}}\) and \(S_{\text{dist}}\) relative to the ground truth label {\(Q^{\text{true}}\)}, thereby adaptively adjusting the confidence in different scores. Specifically, if the memory-based score $S_{\text{dist}}$ is closer to the ground truth, $\alpha_{\text{target}}$ will be larger, indicating higher confidence in the memory score; otherwise, the model places more trust in the reference score $S_{\text{ref}}$.
Additionally, when the reference image is unavailable, the model relies solely on the memory matching score for quality assessment. Owing to the absence of a reference, the weighting parameter $\alpha = 0$ in Equation (\ref{Q_score}) becomes inactive, and the model fully depends on the memory bank for evaluation. This ensures accurate image quality prediction under no-reference conditions. The final quality score is given by $Q = S_{\text{dist}}$.

During the training phase, to reduce the redundancy between the model's memory units and encourage it to learn more independent feature representations, we use a memory loss based on decorrelation constraints to optimize and update the memory bank, expressed as:
\begin{equation}\label{memory_loss}
	\begin{aligned}
\mathcal{L}_{\text{memory}} = \lambda \left( \| \mathbf{C} \|_F - \sum_i \sqrt{\mathbf{C}_{ii}^2 + \epsilon} \right)
	\end{aligned}
\end{equation}
	where \(\lambda \) is a hyperparameter and \(\mathbf{C} \) is the covariance matrix of the features, which represents the correlation between different feature dimensions. \(\| \mathbf{C} \|_F \) denotes the Frobenius norm of the covariance matrix, which measures its overall magnitude. \(\epsilon \) is a small constant to prevent division by zero. \(\mathbf{C}_{ii} \) refers to the diagonal elements of the covariance matrix, indicating the variance of each feature. By minimizing the decorrelation loss, the correlation between different memory units is reduced, allowing each unit to focus on learning distinct distortion patterns. This enhances the model’s ability to differentiate various distortion types, enabling diversified learning among memory units. As a result, the model achieves more precise recognition and distinction of different distortion types, improving the overall effectiveness of the quality assessment framework.
	\begin{table*}[h]
		\renewcommand\arraystretch{0.9}
		\centering
		\caption{Detailed information of the image quality assessment databases.}
		\label{tab:database}
		\setlength\tabcolsep{12pt}
		\begin{tabular}{lccccccccc}
			\Xhline{2.2pt}\hline
			Dataset &LIVE \cite{LIVE} &CSIQ \cite{CSIQMAD}  & TID2013 \cite{TID2013}  & KADID-10K \cite{kadid}  &PIPAL \cite{jinjin2020pipal}\\
			\midrule
			\rowcolor{shadegray}	Reference images &29 & 30&25 &81 &250\\
			Distorted images 	& 779 	& 866 &3,000 &10,125&25,850 \\
			\rowcolor{shadegray}		Distortion types & 5 & 6 & 24 &25&40 \\
			Image resolution &- & 512 $\times$ 512 & 512 $\times$ 384& 512 $\times$ 384& 288 $\times$ 288\\
			\rowcolor{shadegray}		MOS range   &[0, 100] & [0, 1] & [0, 9]& [1, 5] &[868, 1857]\\
			\Xhline{1.4pt}
		\end{tabular}
	\end{table*}	
	\begin{table*}[!t]
		\centering
		\setlength{\tabcolsep}{10pt}
		\renewcommand\arraystretch{1.1}
		\small
		\caption{Performance comparison of the proposed MQAF algorithm against state-of-the-art FR-IQA and NR-IQA algorithms on benchmark datasets. The best, second, and third results are bolded and highlighted in {\lredcell \bf red}, {\lgreencell \bf green}, and {\lbluecell \bf blue}, respectively. Additionally, MQAF-R presents the performance of our method in reference mode, while MQAF-NR shows its performance in no-reference mode.}
			\begin{tabular}{clcccccccc}
				\Xhline{2pt}
				\multirow{2}{*}{Type} &\multirow{2}{*}{Method} & \multicolumn{2}{c}{LIVE \cite{LIVE}} & \multicolumn{2}{c}{CSIQ \cite{CSIQMAD}} & \multicolumn{2}{c}{TID2013 \cite{TID2013}} & \multicolumn{2}{c}{KADID-10K \cite{kadid}} \\
				\cmidrule(lr){3-4} \cmidrule(lr){5-6} \cmidrule(lr){7-8} \cmidrule(lr){9-10}
				& & PLCC & SRCC & PLCC & SRCC & PLCC & SRCC & PLCC & SRCC \\
				\midrule
				
				\multirow{21}{*}{FR} 
				& \gr{PSNR} & \gr{0.791} & \gr{0.807} & \gr{0.781} & \gr{0.801} & \gr{0.663} & \gr{0.686} & \gr{0.667} & \gr{0.672} \\
				& SSIM \cite{wang2004image} & {0.847} & {0.851} & {0.819} & {0.832} & {0.665} & {0.627} & {0.780} & {0.778} \\
				& \gr{MS-SSIM \cite{MS-SSIM}} & \gr{0.886} & \gr{0.903} & \gr{0.864} & \gr{0.879} & \gr{0.785} & \gr{0.729} & \gr{0.835} & \gr{0.834} \\
				& VIF \cite{VIF} & {0.948} & {0.952} & {0.898} & {0.899} & {0.771} & {0.677} & {0.676} & {0.669} \\
				& \gr{MAD \cite{CSIQMAD}} & \gr{0.904} & \gr{0.907} & \gr{0.934} & \gr{0.932} & \gr{0.803} & \gr{0.773} & \gr{0.829} & \gr{0.827} \\
				& FSIM \cite{FSIM} & {0.910} & {0.920} & {0.902} & {0.915} & {0.876} & {0.851} & {0.850} & {0.850} \\
				& \gr{VSI \cite{VSI}} & \gr{0.877} & \gr{0.899} & \gr{0.912} & \gr{0.928} & \gr{0.898} & \gr{0.894} & \gr{0.875} & \gr{0.876} \\
				& GMSD \cite{GMSD} & {0.909} & {0.910} & {0.938} & {0.939} & {0.858} & {0.804} & {0.847} & {0.846} \\
				& \gr{NLPD \cite{NLPD}} & \gr{0.882} & \gr{0.889} & \gr{0.913} & \gr{0.925} & \gr{0.832} & \gr{0.799} & \gr{0.819} & \gr{0.820} \\
				& WaDIQaM-FR \cite{WaDIQaM-FR} & {0.980} & {0.970} & {0.967} & {0.962} & {0.946} & {0.940} &  0.935 &0.931 \\
				& \gr{DeepQA \cite{DeepQA}} & \gr{0.982} &\gr \lbluecell \bf 0.981 & \gr{0.965} & \gr{0.961} & \gr{0.947} & \gr{0.939} & \gr{0.910} & \gr  0.912 \\
				& PieAPP \cite{pieapp} & \lbluecell \bf 0.986 & {0.977} & {0.975} & {0.973} & {0.946} & {0.945} & {0.832} & {0.830} \\
				& \gr{DeepFL-IQA \cite{DeepFL-IQA}} & \gr{0.978} & \gr{0.972} & \gr{0.946} & \gr{0.930} & \gr{0.876} & \gr{0.858} & \gr{0.938} & \gr{0.936} \\
				& JND-SalCAR \cite{JND-SalCAR} & \lgreencell \bf 0.987 &  0.984 & \lbluecell \bf 0.977 & \lbluecell \bf  0.976 &  0.956 &   0.949 & \lbluecell \bf 0.960 & \lbluecell \bf{0.959} \\
				& \gr{LPIPS-VGG \cite{LPIPS}} & \gr{0.978} & \gr{0.972} & \gr{0.970} & \gr{0.967} & \gr{0.944} & \gr{0.936} & \gr- &\gr - \\
				& DISTS \cite{DISTS} & {0.954} & {0.954} & {0.928} & {0.929} & {0.855} & {0.830} & {0.886} & {0.887} \\
				& \gr{A-DISTS \cite{ADISTS}} & \gr{0.955} & \gr{0.955} & \gr{0.947} & \gr{0.941} & \gr{0.858} & \gr{0.835} & \gr{0.892} & \gr{0.892} \\
				& DeepWSD \cite{liao2022deepwsd} & {0.961} & {0.962} & {0.950} & {0.965} & {0.870} & {0.874} & {0.883} & {0.883} \\
				& \gr{TOPIQ-FR \cite{Topiq}} & \gr   0.984 & \gr  \lredcell \bf 0.984 & \gr  \lredcell \bf 0.980 &  \gr \lgreencell \bf 0.978 & \gr  \lbluecell \bf 0.958 & \gr  \lbluecell \bf 0.954 &  \gr - & \gr  - \\
				&\textbf{MQAF-R (ResNet50)}   & 0.982 &  0.980 &\lbluecell \bf 0.977 & \lredcell \bf 0.979 & \lgreencell \bf 0.964 & \lredcell \bf 0.966 &\lgreencell \bf 0.964  &\lgreencell \bf 0.963 \\
				&  \gr \textbf{MQAF-R (VGG16)}   &  \gr \lredcell \bf 0.988 &  \gr \lgreencell \bf 0.983 &  \gr \lgreencell \bf0.978 &  \gr  \lgreencell \bf 0.978 &  \gr  \lredcell \bf 0.965 &  \gr  \lgreencell \bf 0.964 &  \gr  \lredcell \bf 0.966  &  \gr  \lredcell \bf 0.965 \\
				\midrule
				
				\multirow{10}{*}{NR} 
				& WaDIQaM-NR \cite{WaDIQaM-FR} & {0.955} & {0.960} & {0.844} & {0.852} & {0.855} & {0.835} & {0.752} & {0.739} \\
				&  \gr DBCNN \cite{zhang}    &  \gr 0.971  &  \gr 0.968  &  \gr 0.959  &  \gr 0.946   &  \gr 0.865  &  \gr 0.816  &  \gr 0.856  &  \gr 0.851  \\
				&TIQA \cite{you2021transformer}     &0.965 &0.949  &0.838  &0.825  &0.858   &0.846  &0.855  &0.85 \\
				&  \gr MetaIQA \cite{zhu2020metaiqa} & \gr{0.959} & \gr{0.960} & \gr{0.908} & \gr{0.899} & \gr{0.868} & \gr{0.856} & \gr{0.775} & \gr{0.762} \\
				& {SS-IQA \cite{2024}} & {0.980} & \lgreencell \bf {0.978} & {0.969} & {0.960} & {0.910} & {0.891} & {0.895} & {0.896} \\
				&  \gr BIQA, M.D \cite{tip3} & \gr{0.978} & \gr{0.969} & \gr{0.925} & \gr{0.903} & \gr{0.859} & \gr{0.835} & \gr  - & \gr  - \\
				&KGANet \cite{zhou2024multitask} &0.966&0.963&0.963&0.954&{\lbluecell\textbf{0.933}}&{ \lbluecell\textbf{0.927}}&{\lbluecell\textbf{0.943}}&{ \lbluecell\textbf{0.940}} \\
				& \gr QAL-IQA \cite{zhou2025blind} & \gr 0.973 & \gr \lbluecell \bf 0.971 & \gr \lgreencell \bf 0.970 & \gr \lbluecell \bf {0.963}& \gr -& \gr -& \gr 0.910 & \gr  0.908 \\
				&\textbf{MQAF-NR (ResNet50)}   &\lbluecell \bf 0.984 &\lredcell \bf 0.980 &\lredcell \bf 0.971 & \bf \lredcell \bf 0.972 & \lredcell \bf 0.963 & \lredcell \bf 0.960 &\lredcell \bf 0.962  &\lredcell \bf 0.958 \\
				&  \gr  \textbf{MQAF-NR (VGG16)}   &  \gr  \lredcell \bf 0.987 &  \gr  \lredcell \bf 0.980 &  \gr \lbluecell \bf  0.969 &  \gr \lgreencell \bf 0.971 &   \gr  \lgreencell \bf 0.953 &   \gr  \lgreencell \bf 0.951 &  \gr  \lgreencell \bf 0.958  &  \gr  \lgreencell \bf 0.956 \\
				\Xhline{1.4pt}
			\end{tabular}
			\label{tab:comparison}
		\end{table*}	
Finally, in Equation (\ref{LOSS}), we use the mean squared error (MSE) as the quality regression loss function $\mathcal{L}_{\text{pre}}$ and optimize it jointly with the memory loss $\mathcal{L}_{\text{memory}}$ and error-based confidence loss $\mathcal{L}_{\alpha}$ within our proposed framework (MQAF). The overall loss function $\mathcal{L}_{\text{total}}$ is defined as follows:
\begin{equation}
	\mathcal{L}_{\text{total}} = \mathcal{L}_{\text{pre}} + \lambda \cdot \mathcal{L}_{\text{memory}} +  \mathcal{L}_{\alpha}
\end{equation}
Specifically, $\mathcal{L}_{\alpha}$ only updates the parameters of AWN; therefore, no weighting coefficient is needed here.

\subsection{Connection to the HVS and existing methods}
	Existing FR-IQA methods \cite{LPIPS, liao2022deepwsd, liao2024image} predict quality scores by calculating the feature distance between the reference and distorted images to enhance perceptual consistency. However, their evaluation performance heavily relies on high-quality reference images as benchmarks. When reference images suffer from quality defects or are unavailable, distance-based measurement mechanisms fail, leading to evaluation bias. NR-IQA methods \cite{gan2,maeiqa, vcrnet} attempt to generate pseudoreference images through image restoration for guiding evaluation, but their performance is limited by the accuracy and completeness of image reconstruction algorithms. In contrast, the memory-driven quality-aware assessment framework (MQAF) proposed in this paper innovatively builds a memory bank that stores various distortion patterns. It achieves quality assessment by matching the distorted image with prestored distortion pattern memory units in the bank and switches evaluation modes on the basis of the availability of reference images to reduce dependency. The MQAF we propose simulates the memory-driven mechanism of the human brain for visual information, not only breaking through the rigid dependency on reference images in traditional methods but also offering potential for extending its memory-driven architecture to broader quality assessment tasks.

\section{Experiments}\label{Experiments}
	\subsection{Experimental setups}
	
\textit{1) Databases}: We evaluated the performance of our proposed method via five publicly available datasets, including the LIVE \cite{LIVE}, CSIQ \cite{CSIQMAD}, TID2013 \cite{TID2013}, KADID-10k \cite{kadid}, and PIPAL \cite{jinjin2020pipal} datasets. These datasets vary significantly in terms of size, distortion type, and subjective score annotations, as summarized in Table \ref{tab:database}.

\textit{2) Evaluation metrics}: For evaluation metrics, we selected the Spearman rank order correlation coefficient ({SRCC}) \cite{SROCC} and the Pearson linear correlation coefficient (PLCC) \cite{PLCC} to compare the performance of our proposed memory-driven and dynamic quality-aware assessment framework (MQAF) with that of other FR-IQA methods.
The SRCC and PLCC can be formulated as follows:
\begin{equation}
	PLCC  = \frac{\sum_{i=1}^n (X_i - \bar{X} ) (S_i-\bar{S} )} {\sqrt{\sum_{i=1}^n (X_i-\bar{X} )^2} \sqrt{\sum_{i=1}^n (S_i-\bar{S} )^2}} 
\end{equation}
where $X$ and $S$ denote the subjective and objective qualities, respectively, and
\begin{equation}
	SRCC = 1-\frac{6 \sum_{i=1}^n d_i^2}{n\left(n^2-1\right)}
\end{equation}
where $n$ is the total number of samples and where $d_i$ indicates the rank difference between the subjective and predicted scores for the $i$-th sample.

\textit{3) Implementation details}: All experiments in this study were conducted via the PyTorch framework on an NVIDIA GeForce RTX 4090 GPU. During the image preprocessing stage, all the input images were randomly cropped to a size of 224$\times$224$\times$3 before being fed into the MQAF model. During training, the main hyperparameters were configured as follows: the learning rate was set to $8\times10^{-5}$, the weight decay was $1\times10^{-5}$, the batch size was 16, and the total number of training epochs was 200. The Adam optimizer was employed to ensure efficient and stable parameter updates.

\subsection{Experimental Results}
\subsubsection{Comparison with the State-of-the-Art Method}
	In our study, to comprehensively evaluate the effectiveness of the proposed MQAF framework, we conducted performance comparisons against various classic 27 IQA methods on standard datasets.
These FR-IQA methods include PSNR, SSIM \cite{wang2004image}, MS-SSIM \cite{MS-SSIM}, VSI \cite{VSI}, VIF \cite{VIF}, FSIM \cite{FSIM}, GMSD \cite{GMSD}, NLPD \cite{NLPD}, WaDIQaM-FR \cite{WaDIQaM-FR}, PieAPP \cite{pieapp}, MAD \cite{CSIQMAD}, DISTS \cite{DISTS}, A-DISTS \cite{ADISTS}, DeepWSD \cite{liao2022deepwsd}, DeepQA \cite{DeepQA}, DeepFL-IQA \cite{DeepFL-IQA}, JND-SalCAR \cite{JND-SalCAR}, LPIPS-VGG \cite{LPIPS}, and TOPIQ \cite{Topiq}. Additionally, we included popular NR-IQA methods, including WaDIQaM-NR \cite{wadiqam-nr}, DBCNN \cite{zhang}, MetaIQA \cite{zhu2020metaiqa}, TIQA \cite{you2021transformer}, BIQA, M.D \cite{tip3}, SS-IQA \cite{2024}, KGANet \cite{zhou2024multitask}, and QAL-IQA \cite{zhou2025blind}.

\begin{table}[!t]
	\centering
	\small
	\renewcommand{\arraystretch}{1.1}
	\setlength{\tabcolsep}{20pt}
	\caption{SRCC results of MQAF-R on the PIPAL dataset. The best results are shown in \textbf{bold}.}
	\begin{tabular}{lccc}
		\Xhline{2.2pt}\hline
		\multirow{2}{*}{{Methods}}  & \multicolumn{2}{c}{{PIPAL Validation \cite{jinjin2020pipal}}} \\
		\cline{2-3}
		
		& PLCC & SRCC  \\
		\Xhline{0.6pt}
		\rowcolor{shadegray}	PSNR & 0.289 & 0.255  \\
		SSIM \cite{wang2004image} & 0.409 & 0.363  \\
		\rowcolor{shadegray}	MS-SSIM \cite{MS-SSIM} & 0.566 & 0.491  \\
		FSIM \cite{FSIM} & 0.562 & 0.468  \\
		\rowcolor{shadegray}	VSI \cite{VSI} & 0.515 & 0.450  \\
		VIF \cite{VIF}& 0.572 & 0.503\\
		\rowcolor{shadegray}	NLPD \cite{NLPD} & 0.558 & 0.370  \\
		GMSD \cite{GMSD} & 0.656 & 0.585  \\
		\rowcolor{shadegray}	MAD \cite{MAD} & {0.702} & 0.649  \\
		\Xhline{0.5pt}
		PieAPP \cite{pieapp} & 0.696 & {0.704} \\
		\rowcolor{shadegray}	LPIPS-Alex \cite{LPIPS} & 0.606 & {0.569}  \\
		LPIPS-VGG \cite{LPIPS} & 0.611 & {0.551}  \\
		\rowcolor{shadegray}	DISTS \cite{DISTS} & {0.634} & 0.608 \\
		\Xhline{0.5pt}
		MQAF-R (Ours) &\bf 0.763  & \bf0.776  \\
		\hline\hline\Xhline{1.4pt}
	\end{tabular}
	\label{tab:PIPAL_srcc}
\end{table}

\begin{table}[h]
	\centering
	\small
	\setlength{\tabcolsep}{2pt}
	\renewcommand{\arraystretch}{1.3}
	\caption{Comparison of cross-dataset performance on KADID-10K \cite{kadid} dataset. The best results are highlighted in \textbf{bold}.}
	\begin{tabular}{lcccccccccccc}
		\Xhline{2.2pt}\hline
		\multirow{4}{*}{{Method}} & 
		\multicolumn{6}{c}{{Training Set: KADID-10K}} & \\
		\cmidrule{2-7}
		& \multicolumn{2}{c}{LIVE } & \multicolumn{2}{c}{CSIQ } & \multicolumn{2}{c}{TID2013 } \\
		\cmidrule(lr){2-3}\cmidrule(lr){4-5} \cmidrule(lr){6-7} 
		& PLCC & SRCC & PLCC & SRCC & PLCC & SRCC \\
		\Xhline{0.6pt}
		\rowcolor{shadegray}	PSNR &0.872&0.0.876&0.796&0.806&0.702&0.639\\
		WaDIQaM-FR \cite{WaDIQaM-FR}  & 0.837 & 0.883   & - & -  & 0.741 & 0.698 \\
		\rowcolor{shadegray}	RADN \cite{RADN} & 0.878 & 0.905   & - & -  & 0.796 & 0.747 \\
		PieAPP \cite{pieapp} &0.910  &0.918  &0.890   &0.897  &0.831   &0.844   \\
		\rowcolor{shadegray}	LPIPS-Alex \cite{LPIPS}&0.900  &0.922  &0.893  &\bf0.928  &0.787  &0.776\\
		MQAF-R (Ours)    &\bf  0.913 & \bf0.941 & \bf 0.903 & 0.910 & \bf0.845 & \bf 0.847 \\
		\hline\hline\Xhline{1.4pt}
	\end{tabular}
	\label{tab:cross}
\end{table}
\begin{table*}[h]
	\small 
	\setlength{\tabcolsep}{9pt}
	\centering
	\renewcommand\arraystretch{1.1}
	\caption{SROCC results for individual distortion types on the LIVE\cite{LIVE} and CSIQ \cite {CSIQMAD} databases. \textbf{The best result is shown in bold} and the second-best result is \underline{underlined}.}
	\begin{tabular}{lccccccccccc}
		\Xhline{2.2pt}\hline
		\multirow{2}{*}{Method} & \multicolumn{5}{c}{LIVE} & \multicolumn{6}{c}{CSIQ} \\
		\cmidrule(lr){2-6}\cmidrule(lr){7-12}
		& JP2K & JPEG & WN & GB & FF  & WN & JPEG & JP2K  & PN & GB & CC \\
		\midrule
		\rowcolor{shadegray}	PSNR & 0.895 & 0.881 & 0.985 & 0.782 & 0.891 & \underline{0.963} & 0.888 & 0.936 & 0.934 & 0.929 & 0.862 \\
		SSIM \cite{wang2004image}    & 0.961 & 0.972 & 0.969 & 0.952 & 0.956 & 0.897 & 0.956 & 0.961 & 0.892 & 0.961 & 0.792 \\
		\rowcolor{shadegray}	 VIF  \cite{VIF}  & 0.969 & \bf0.984 & 0.985 & 0.972 & 0.965 & 0.958 & \underline{0.971} & 0.967 & \underline{0.951} & \underline{0.975}    & 0.935     \\
		GSMD  \cite{GMSD}      & 0.968 & 0.973 & 0.974 & 0.957 & 0.942 &0.944 & 0.963 & 0.965 & 0.939 & 0.959 & 0.935 \\
		\rowcolor{shadegray}		FSIMc   \cite{FSIM}      & \underline{0.972} & 0.979 & 0.971 & 0.968 & 0.950 & 0.936 & 0.966 & 0.970 & 0.937 & 0.973 & 0.944 \\
		IFC \cite{IFC} & 0.910 & 0.944 & 0.937 & 0.965 & 0.964 & 0.846 & 0.940 & 0.926 & 0.828 & 0.959 & 0.542 \\
		\rowcolor{shadegray}		DeepQA   \cite{DeepQA}   & \underline{0.972} & 0.980 & 0.986 & \bf0.982 & 0.963 & 0.904 & 0.948 & \underline{0.972} & 0.930 & 0.970 & \underline{0.956} \\
		DRF-IQA   \cite{kim2020dynamic}     & \bf0.978 & \underline{0.982} & \bf0.988 & \bf0.982 &\underline{0.967} & 0.910 &0.961 & 0.968 & 0.942 & 0.972& \bf0.963 \\
		\rowcolor{shadegray}			MQAF-R (Ours)& 0.968 & 0.970 & \underline{0.987} & \underline{0.980} & \bf0.975  &\bf0.983&\bf0.972  &\bf0.983 &\bf0.972 &\bf0.987 &0.950  \\
		\hline\hline\Xhline{1.4pt}
	\end{tabular}
	\label{tab:individual_types}
\end{table*}
Table \ref{tab:comparison} compares the performance of the proposed MQAF method with that of state-of-the-art full-reference (FR-IQA) and no-reference (NR-IQA) image quality assessment approaches across four public datasets. In particular, we evaluate MQAF under two operational modes. In the reference mode (MQAF-R), the model performs quality prediction by incorporating both the reference image and a memory matching mechanism. In the no-reference mode (MQAF-NR), the model relies entirely on internal memory to assess image quality. The bold entries in Table \ref{tab:comparison} highlight the results for these two modes. The experimental results demonstrate that MQAF consistently delivers strong performance regardless of the availability of a reference image. On the KADID-10K dataset, it achieves the best results and significantly outperforms existing methods. In the reference mode, MQAF attains a PLCC score on the CSIQ dataset that is only 0.002 lower than the best result achieved by TOPIQ-FR \cite{Topiq}, indicating a minimal gap. Notably, under the no-reference setting, MQAF still achieves high PLCC and SRCC scores, surpassing existing mainstream NR-IQA methods. This further confirms the effectiveness and robustness of the proposed memory mechanism in no-reference scenarios. Additionally, we compare the performance of MQAF under both modes when it is combined with different backbone networks. The results show that, whether in MQAF-R or MQAF-NR mode, the use of VGG16 or RseNet50 as the backbone leads to optimal or near-optimal performance across all datasets. This finding further confirms the advantages and versatility of our memory-based design in IQA tasks.
\begin{table*}[!t] 
	\setlength{\tabcolsep}{6.5pt}
	\centering
	\renewcommand\arraystretch{1.2}
	\small
	\caption{SROCC value comparison in leave–one-distortion-out cross-validation on the TID2013 dataset. The best result is shown in \textbf{bold}, the second-best result is \underline{underlined}, and $^\dagger$ denotes the reference mode (MQAF-R) and no-reference mode model (MQAF-NR), whereas $*$ indicates the best results under the MQAF-R model.}
	\begin{tabular}{lccccccccccc}
		\Xhline{2.2pt}\hline
		Dist. type  & \makecell{BRISQUE \\ \cite{zhang2015feature}} & \makecell{ILNIQE \\ \cite{mittal2012no}} & \makecell{CORNIA \\ \cite{ye2012unsupervised}} & \makecell{HOSA \\ \cite{xu2016blind}} & \makecell{WaDIQaM-NR \\ \cite{WaDIQaM-FR}} & \makecell{DBCNN \\ \cite{zhang}} & \makecell{MetalQA\\ \cite{zhu2020metaiqa}} & \makecell{$^\dagger$MQAF-NR \\ (Ours)} &  \makecell{$^\dagger$MQAF-R\\ (Ours)}\\
		\midrule
		\rowcolor{shadegray}AGN  & 0.9356 & 0.8760 & 0.4465 & 0.7582 & 0.9080 & 0.7585 & \underline{0.9473} & \bf0.9650 & *\bf0.9694\\
		ANC  & 0.8114 & 0.8159 & 0.1020 & 0.4670 & 0.8700 & 0.8423 & \underline{0.9240} & \bf0.9450 & 0.9188\\
		\rowcolor{shadegray}	SCN  & 0.5457 & 0.9233 & 0.6697 & 0.6246 & 0.8802 & 0.7800 & \underline{0.9534} & \bf0.9758 & 0.9726\\
		MN   & 0.5852 & 0.5120 & 0.6096 & 0.5125 & 0.8065 & 0.4501 & \underline{0.7277} & \bf0.8362 & *\bf0.8831\\
		\rowcolor{shadegray}	HFN  & 0.8956 & 0.8685 & 0.8402 & 0.8285 & 0.9314 & 0.8192 & \bf{0.9518} & \underline{0.9509} &*\bf0.9667\\
		IN   & 0.6559 & 0.7551 & 0.3526 & 0.1889 & 0.8779 & 0.5798 & \bf{0.8653} & \underline{0.8384} & 0.8172\\
		\rowcolor{shadegray}	QN   & 0.6555 & 0.8730 & 0.3723 & 0.4145 & 0.8541 & 0.7502 & \underline{0.7454} & \bf0.9078 & 0.8917\\
		GB   & 0.8656 & 0.8142 & 0.8879 & 0.7823 & 0.7520 & 0.9015 & \bf0.9767 & \underline{0.9730} & *\bf0.9836\\
		\rowcolor{shadegray}	DEN  & 0.6143 & 0.7500 & 0.6475 & 0.5436 & 0.7680 & 0.8103 & \underline{0.9383} & \bf0.9621 & *\bf0.9704\\
		JPEG & 0.5186 & 0.8349 & 0.8295 & 0.8318 & 0.7841 & 0.8131 & \underline{0.9340} & \bf0.9566 & *\bf0.9575\\
		\rowcolor{shadegray}	JP2K & 0.7592 & 0.8578 & 0.8611 & 0.5097 & 0.8706 & 0.8162 & \underline{0.9586} & \bf0.9707 & *\bf0.9803\\
		JGTE & 0.5604 & 0.2827 & 0.7282 & 0.4494 & 0.5191 & 0.7023 & \bf0.9297 & \underline{0.9040} & *\bf0.9322\\
		\rowcolor{shadegray}	J2TE & 0.7003 & 0.5248 & 0.4817 & 0.1405 & 0.4322 & 0.6577 & \bf0.9034 & \underline{0.9024} & *\bf0.9286\\
		NEPN & 0.3111 & -0.0805 & 0.3471 & 0.2163 & 0.1230 & 0.3774 & \bf0.7238 & \underline{0.5142} & *\bf0.7715\\
		\rowcolor{shadegray}	Block & 0.2659 & -0.1357 & 0.2345 & 0.3767 & 0.4059 &\underline{0.4169}  & {0.3899} & \bf0.5331 & *\bf0.7086\\
		MS   & 0.1852 & 0.1845 & 0.1775 & 0.0633 & \bf0.4596 & 0.1023 &\underline{0.4016}  & 0.3822 & *\bf0.7277\\
		\rowcolor{shadegray}	CTC  & 0.0182 & 0.0141 & 0.2122 & 0.0466 & 0.5401 & 0.2946 & \underline{0.7637} & \bf0.8016 & 0.7725\\
		CCS  & 0.2142 & -0.1628 & 0.2299 & -0.1390 & 0.5640 & 0.3423 & \bf{0.8294} & \underline{0.8067} & *\bf0.8629\\
		\rowcolor{shadegray}	MGN  & 0.8777 & 0.6932 & 0.4931 & 0.5491 & 0.8810 & 0.6700 & \underline{0.9392} & \bf0.9533 & *\bf0.9573\\
		CN   & 0.4706 & 0.3599 & 0.5069 & 0.3740 & 0.6466 & 0.5544 & \bf0.9516 & \underline{0.9514} & *\bf0.9570\\
		\rowcolor{shadegray}	LCNI & 0.8238 & 0.8287 & 0.7191 & 0.5053 & 0.6882 & 0.7600 & \bf0.9779 & \underline{0.9743} & 0.9743\\
		ICQD & 0.4883 & 0.7487 & 0.7757 & 0.8036 & 0.7965 & 0.7608 & \underline{0.8597} & \bf0.9275 &*\bf0.9416\\
		\rowcolor{shadegray}	CHA  & 0.7470 & 0.6793 & 0.6937 & 0.6657 & 0.7950 & 0.7097 & \bf0.9269 &\underline{0.9125}  & *\bf0.9275\\
		SSR  & 0.7727 & 0.8650 & 0.8867 & 0.8273 & 0.8220 & 0.7700 & \bf0.9744 & \underline{0.9718} & *\bf0.9753\\
		\hline\hline\Xhline{1.4pt}
	\end{tabular}
	\label{crosstypes}
\end{table*}
\begin{table*}[!t]
	\setlength{\tabcolsep}{9.2pt}
	\renewcommand\arraystretch{1}
	\centering
	\small
	\caption{Comparison of computational complexity on different methods and backbone networks.}
	\begin{tabular}{lccccccccccc}
		\Xhline{2.2pt}\hline
		\textbf{NAME} & PieAPP \cite{pieapp}& LPIPS \cite{LPIPS}& DISTS \cite{DISTS} & PDL \cite{delbracio2021projected}& WaDIQaM-FR \cite{WaDIQaM-FR}& A-DISTS \cite{ADISTS} \\
		\midrule
		\rowcolor{shadegray}	MACs / G & 30.25 & 20.04 & 20.40  & 25.40 & 20.04 & 20.04  \\
		Params / M & 68.38 & \bf14.71 & 16.02  & 20.02 & \underline{14.72} & \bf14.71  \\
		\rowcolor{shadegray}	Runtime / min & 29.74 & \bf1.98 & 3.66  & 11.66 & 15.43 & 3.10  \\
		\midrule
		\textbf{NAME}  & AHIQ \cite{AHIQ}& DeepWSD \cite{liao2022deepwsd}& TOPIQ \cite{Topiq} & DeepDC \cite{zhu2022deepdc}& \makecell{MQAF-R \\w/VGG16} & \makecell{MQAF-R \\w/ResNet50}  \\
		\midrule
		\rowcolor{shadegray}	MACs / G & 1689.91 & 25.40& \bf7.80 & 19.05 &30.70& \underline{8.27}   \\
		Params / M  & 133.39 & 20.02& 32.35 & 26.54 &16.29  & 28.22   \\
		\rowcolor{shadegray}		Runtime / min & 60.24 & 11.66& 13.50 & 3.47 & \underline{2.31} & 2.57  \\
		\hline\hline\Xhline{1.4pt}
	\end{tabular}
	\label{params}
\end{table*}

In addition, we further evaluated the proposed MQAF method on the large-scale PIPAL \cite{jinjin2020pipal} dataset, which contains 40 types of complex distortions, and compared it with several state-of-the-art FR-IQA methods. The experiments were conducted strictly on the basis of the official training and validation split provided by the PIPAL dataset to ensure fairness and reproducibility. The results are shown in Table \ref{tab:PIPAL_srcc}. It can be observed that MQAF demonstrates stable performance under various distortion conditions and outperforms mainstream FR-IQA methods in overall metrics, further validating the effectiveness and superiority of the proposed method in complex and realistic scenarios.

\subsubsection{Performance Evaluation Cross Different Databases}
To evaluate the generalization ability of our method on the reference-based (MQAF-R) dataset, we conducted cross-database experiments without any fine-tuning or parameter adaptation. The model was trained on the KADID-10K dataset and tested on the LIVE, CSIQ, and TID2013 datasets. Table \ref{tab:cross} presents the comparison results between our method and representative approaches, including PSNR, WaDIQaM-FR \cite{WaDIQaM-FR}, RADN \cite{RADN}, PieAPP \cite{pieapp}, and LPIPS-Alex \cite{LPIPS}. The experimental results show that our method achieves superior performance when LIVE and TID2013 are used as test datasets. Although the SRCC metric on the CSIQ dataset does not reach the highest score, our method still maintains a leading performance in terms of PLCC, further demonstrating the strong generalization ability and robustness of MQAF-R in handling complex distortions.

\subsubsection{Performance Evaluation on Individual Distortion Types}
To evaluate the performance of the proposed method on individual distortion types, we conducted experiments on seven types of distortions from the LIVE and CSIQ datasets, including JPEG2000 compression (JP2K), JPEG compression (JPEG), white noise (WN), Gaussian blur (GBLUR), fast fading distortion (FF), Poisson noise (PN), and contrast distortion (CC). The experimental results are shown in Table \ref{tab:individual_types}. As illustrated in the results, our method demonstrates superior performance across most distortion types, with particularly strong results on the CSIQ dataset. For the LIVE dataset, even in cases where our method did not achieve the best score, it consistently ranked among the top two methods, indicating its robustness and adaptability across diverse distortion scenarios.
\begin{table*}[t]
	\centering
	\setlength{\tabcolsep}{11.5pt}
	\renewcommand\arraystretch{1}
	\caption{Results of ablation results on four benchmark datasets. The best result is shown in \textbf{bold}.}
	\begin{tabular}{lcccccccccc}
		\Xhline{2.2pt}\hline
		\multirow{2}{*}{Method} & \multicolumn{2}{c}{LIVE} & \multicolumn{2}{c}{CSIQ} & \multicolumn{2}{c}{TID2013} & \multicolumn{2}{c}{KADID-10K} \\
		\cmidrule(lr){2-3} \cmidrule(lr){4-5} \cmidrule(lr){6-7} \cmidrule(lr){8-9}
		& PLCC  & SRCC  & PLCC  & SRCC  & PLCC  & SRCC  & PLCC  & SRCC  \\
		\hline
		\rowcolor{shadegray}	Memory-driven (w/o)   &  0.980 &  0.977   & 0.964 & 0.970  & 0.931 & 0.937 & 0.950 & 0.950 \\
		MQAF-R (Ours)              & 0.982 &  \bf0.980 & \bf 0.977 & \bf 0.979 & \bf 0.964 & \bf 0.966 & \bf 0.964  & \bf 0.963 \\
		\rowcolor{shadegray}	MQAF-NR (Ours)   & \bf 0.984 & \bf 0.980 &0.971 & 0.972 &  0.963 &   0.960 &  0.962  &  0.958 \\
		\hline\hline\Xhline{1.4pt}
	\end{tabular}
	\label{tab:ablation1}
\end{table*}
\subsubsection{Performance Comparison of Cross Distortion Types}
To further evaluate the generalization ability of the proposed model across different distortion types, we conducted experiments via the leave-one-distortion-out cross-validation (LODO-CV) strategy. Specifically, on the basis of the 24 distortion types in the TID2013 dataset, in each iteration, one distortion type was held out as the test set, while the remaining 23 types were used for training. This process was repeated 24 times to ensure that each distortion type was evaluated individually, allowing for a comprehensive assessment of the model’s ability to handle previously unseen distortions. Table \ref{crosstypes} presents the performance of our proposed model under both the no-reference mode (MQAF-NR) and the reference mode (MQAF-R), with each distortion type used as the test set. The results demonstrate that MQAF-NR outperforms existing NR-IQA methods on most distortion types, indicating strong generalizability across distortions. Although the model does not always achieve the top performance on every single type, it consistently ranks among the top two. When operating in reference mode, which incorporates reference image information and a memory-based matching mechanism, the model shows a significant performance boost and achieves consistently superior results across all distortion types.	
\begin{table}[t]	
	\setlength{\tabcolsep}{15pt}
	\renewcommand\arraystretch{1.2}
	\centering
	\small
	\caption{Effect of the number of memory units on model performance on the TID2013 \cite{TID2013} dataset.}
	\label{tab:memory_units}
	\begin{tabular}{cccc}
		\Xhline{2.2pt}\hline
		\multicolumn{1}{c}{Memory Units}	 & 	\textbf{\(\mathbf{\lambda}\)}&PLCC & SRCC \\
		\hline
		\rowcolor{shadegray}	128 & 0.01&0.963&0.962\\
		\bf256 &\bf 0.01 & \textbf{0.964}  & \textbf{0.964}  \\
		\rowcolor{shadegray}	512 & 0.01 & 0.963  & 0.964  \\
		1024 & 0.01 & 0.962  & 0.962  \\
		\rowcolor{shadegray}	2048& 0.01 &0.962 &0.959\\
		\hline\hline\Xhline{1.4pt}
	\end{tabular}
	\label{tab:ablation2}
\end{table}

\begin{table}[t]
	\setlength{\tabcolsep}{15pt}
	\renewcommand\arraystretch{1.1}
	\centering
	\small
	\caption{Sensitivity analysis of the memory loss in Equation (\ref{memory_loss}) hyperparameter \( \lambda \) on the TID2013 \cite{TID2013} dataset.}
	\label{tab:memory_loss}
	\begin{tabular}{lccc}
		\Xhline{2.2pt}\hline
		\textbf{\(\mathbf{\lambda}\)}	&Memory Units  & {PLCC} & {SRCC} \\
		\hline
		\rowcolor{shadegray}	0    & 256   &0.963 & 0.962  \\
		0.001 & 256 &0.964 & 0.964  \\
		\rowcolor{shadegray}	0.005  & 256 & 0.963  & 0.963  \\
		0.01  & 256  &0.960  & 0.963  \\
		\rowcolor{shadegray}	0.05  & 256  & 0.963  & 0.964  \\
		\bf	0.1  &\bf 256  & \bf0.964  &\bf 0.966  \\
		\rowcolor{shadegray}		0.5  & 256  & 0.962  & 0.963  \\
		1  & 256  & 0.962  & 0.962  \\
		\hline\hline\Xhline{1.4pt}
	\end{tabular}
	\label{tab:ablation3}
\end{table}		

\subsubsection{Comparison of Computational Complexity}
To comprehensively evaluate the computational efficiency of different methods in practical applications, we compare the number of parameters (Params/M), the number of multiply accumulate operations (MACs/G), and the runtime (Runtime/min) required to process the entire PIPAL dataset during inference. Table \ref{params} summarizes the complexity performance of several mainstream FR-IQA methods alongside our proposed MQAF-R model with different backbone configurations. The results show that, compared with methods such as AHIQ \cite{AHIQ} and PieAPP \cite{pieapp}, the proposed MQAF-R model demonstrates good computational efficiency under the reference setting with both backbone options. When ResNet50 is used, although its parameter count is slightly higher than that of VGG16 (11.94M), its computational cost measured in MACs is significantly lower at 8.27G, which makes it more efficient than VGG16 and many other baseline models. Both backbone configurations maintain a well-balanced runtime, highlighting MQAF-R’s effective trade-off between accuracy and computational cost.
	\begin{figure*}[!t]
		\centering
		\setlength{\abovecaptionskip}{0.cm}
		\includegraphics[width=0.9\textwidth]{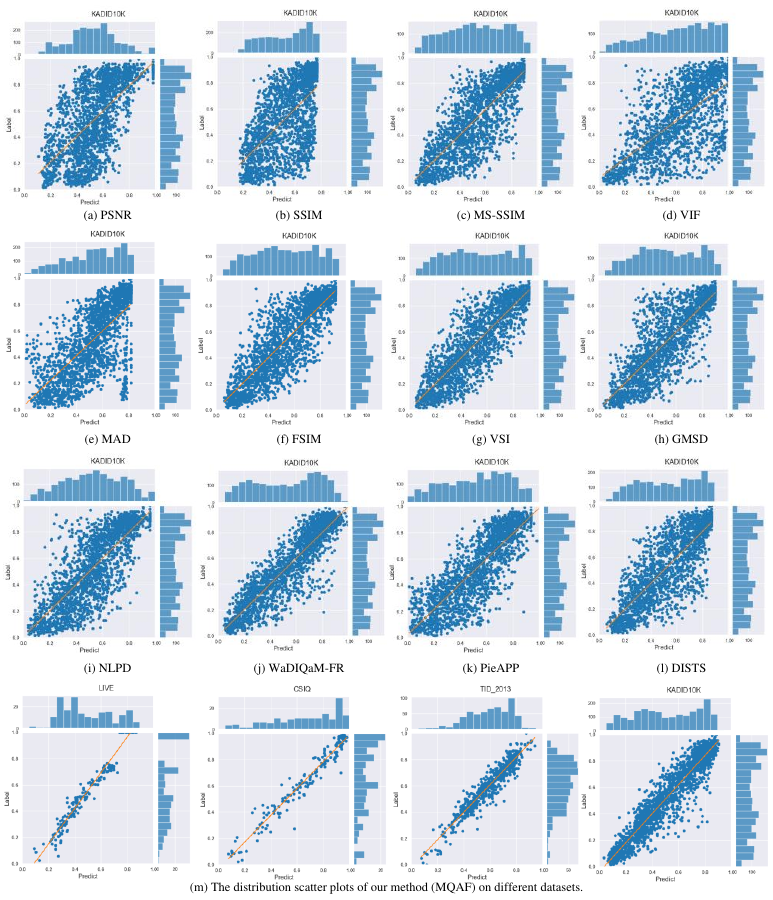}
		\vspace{0.2em}
		\caption{{Scatter plot of prediction results for different FR-IQA methods and our method (MQAF-R) on the KADID-10K \cite{kadid} dataset. The vertical axis represents the ground truth values, whereas the horizontal axis represents the predicted values.}}
		\label{visual2}
	\end{figure*}

\begin{figure*}[!t]
	\centering
	\setlength{\abovecaptionskip}{0.cm}
	\includegraphics[width=0.9\textwidth]{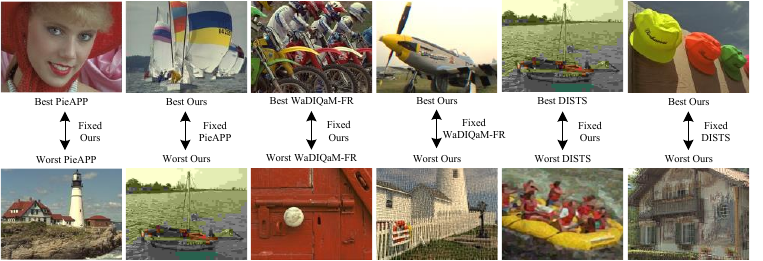}
	\vspace{0.2em}
	\caption{{Comparison of gMAD competition results on the TID2013 dataset among our proposed MQAF-R method, PieAPP \cite{pieapp}, WaDIQaM-FR \cite{WaDIQaM-FR}, and DISTS \cite{DISTS} is presented.}}
	\label{gMAD}
\end{figure*}
\subsection{Ablation Study and Analysis}
To validate the impact of our proposed method MQAF on overall performance, we designed and conducted ablation experiments. All experimental settings remained consistent with the complete model to ensure a fair comparison. Specifically, we first removed the memory mechanism to evaluate the contribution of memory storage and matching in quality assessment. As shown in Table \ref{tab:ablation1}, after eliminating the memory-driven distortion pattern matching, the PLCC and {SRCC} metrics of the model significantly declined across all datasets. Moreover, the memory-driven model alone achieved the best results on the LIVE and TID2013 datasets. These findings confirm the effectiveness of storing and matching various distortion patterns in the memory bank, demonstrating that this mechanism significantly enhances the model's ability to recognize different distortion types, thereby improving overall evaluation performance.

\subsection{Sensitivity Parameter Analysis}
	To further analyse the impact of the number of memory units on model performance, we conducted sensitivity experiments with memory unit counts set to 128, 256, 512, 1024, and 2048 on the TID2013 dataset. The experimental results are presented in Table \ref{tab:ablation2}. As shown in the results, variations in the number of memory units affected both the PLCC and SRCC metrics. When the number of memory units was set to {256}, the model achieved the best performance across all datasets, indicating that an appropriately sized memory bank can effectively enhance the accuracy of quality assessment. Additionally, to validate the effectiveness of the memory loss (decorrelation loss) during training and explore the sensitivity of the hyperparameter \(\lambda \), we conducted further experiments. Table \ref{tab:ablation3} presents the model’s performance under different values of \(\lambda \) (\(0, 0.001, 0.005, 0.01, 0.05, 0.1, 0.5, 1\)). When the decorrelation loss is removed, the model exhibits relatively balanced performance. However, when \(\lambda = 0.1 \), the model achieves the optimal SRCC score. This finding suggests that memory loss plays a role in reducing redundancy among memory units and promoting the learning of more independent feature representations, thereby improving the overall quality assessment performance. All the experimental results in this paper are obtained with the number of memory units set to 256 and $\lambda = 0.1$.

\subsection{Qualitative Evaluation}
	To better illustrate the performance advantages of our proposed method (MQAF), we conducted a visualization analysis. As shown in Figure \ref{radar}, we plotted radar charts comparing MQAF-R with state-of-the-art IQA methods across different datasets. The figure clearly shows that MQAF-R has significant performance advantages across multiple dimensions. Additionally, we generated scatter plots that compare the actual scores and predicted scores of MQAF-R with those of 12 classical FR-IQA methods, as shown in Figure \ref{visual2}. This figure presents the prediction results of different methods on the KADID-10K dataset. The experimental results indicate that the predicted scores of MQAF-R are highly correlated with the ground truth scores, further confirming that MQAF-R achieves outstanding prediction performance on datasets containing various types of distortions. In addition, we employed the group maximum differentiation (gMAD) competition method to evaluate the robustness and discriminative capability of the proposed MQAF-R method. Figure \ref{gMAD} presents the subjective results of MQAF-R compared with those of PieAPP \cite{pieapp}, WaDIQaM-FR \cite{WaDIQaM-FR}, and DISTS \cite{DISTS}. In the competition, when MQAF-R acts as the defender (considering image pairs to be of similar quality) and the other three methods act as attackers, an attack is deemed successful if the attacker finds a counterexample (i.e., a pair with a significant quality difference); otherwise, it is considered a failure. As shown in Figure \ref{gMAD}, MQAF-R, as the defender, accurately identifies image pairs of similar quality, which is consistent with subjective perception, whereas the other methods mistakenly judge these pairs as having significant differences. Conversely, when MQAF-R serves as the attacker, it effectively detects image pairs with large quality differences (counterexamples). Compared with the other three methods, MQAF-R more accurately distinguishes between high-quality and low-quality image pairs, demonstrating superior discriminative ability.
	
\section{Conclusion}\label{Conclusion}
	This paper addresses the limitations of existing image quality assessment methods that heavily depend on high-quality reference images and proposes a memory-driven quality-aware framework (MQAF). By constructing a trainable memory bank with dedicated units for distortion patterns, MQAF autonomously learns and stores typical distortion features, reducing reliance on reference images. When available, reference images are combined with stored distortion patterns for adaptive evaluation; otherwise, quality assessment relies solely on memory matching. This dual-mode operation enhances adaptability across diverse application scenarios. Additionally, an end-to-end joint optimization strategy integrates memory matching loss and quality regression loss for unified training. Even without a reference or with low-quality references, MQAF maintains stable performance, demonstrating strong robustness and broad applicability.

{
	\small
	\bibliographystyle{IEEEtran}
	\bibliography{main}
}

\end{document}